\documentclass{article}
\usepackage{spconf,color,dsfont,bm,epsfig,graphicx,cite,algorithmicx,algorithm}
\usepackage{amsmath,amsfonts,amsthm,amssymb} 
\usepackage{stfloats}

\newtheorem{pro}{Proposition}

\long\def\symbolfootnote[#1]#2{\begingroup
\def\thefootnote{\fnsymbol{footnote}}
\footnote[#1]{#2}\endgroup}

\ninept

\begin{document}

\title{Online Optimal Task Offloading with One-Bit Feedback}
\name{Shangshu Zhao, Zhaowei Zhu, Fuqian Yang, and Xiliang Luo}
\address{ShanghaiTech University, Shanghai, China\\
Email:{ {\{zhaoshsh, zhuzhw, yangfq, luoxl\}}@shanghaitech.edu.cn}}

\maketitle

\begin{abstract}
Task offloading is an emerging technology in fog-enabled networks.
It allows users to transmit tasks to neighbor fog nodes so as to utilize the
computing resources of the networks. In this paper, we investigate a stochastic
task offloading model and propose a multi-armed bandit framework to formulate
this model. We consider the fact that different helper nodes prefer different
kinds of tasks. Further, we assume each helper node just feeds back one-bit
information to the task node to indicate the level of happiness.
The key challenge of this problem lies in the exploration-exploitation tradeoff.
We thus implement a UCB-type algorithm to maximize the
long-term happiness metric. Numerical simulations are given in the end of the
paper to corroborate our strategy.
\end{abstract}


\vspace{-5mm}
\section{Introduction}
\vspace{-2mm}

With the arriving of Internet of Things (IoT), 5G wireless systems, and the embedded
artificial intelligence, more and more data processing capability is required in
mobile devices \cite{2015_Chiang_Fog_IoT}. To benefit from all the available
computational resources, fog computing (or mobile edge computing) has been considered
to be a potential solution to enable computation-intensive and latency-critical
applications at the battery-empowered mobile devices \cite{2017_Dinh_MobileEdge}.

Fog computing promises dramatic reduction in latency and mobile energy consumption
by offloading computation tasks \cite{Mao2017}. Recently, many works have been carried
out addressing the task offloading in fog computing
\cite{2017_Pu_D2D_fog,2017_Yang_DEBTS,2017_Mao_MobileEdge,2015_Kwak_JSAC_DREAM,
2017_You_MobileEdge,2017_Yang_MEETS}. Among these works, some considered the energy
issues and formulated the task offloading as deterministic optimization problems
\cite{2017_You_MobileEdge,2017_Yang_MEETS}. Considering the real-time states of
users and servers, the task offloading problem is a typical stochastic optimization
problem. To make this problem tractable, in
\cite{2017_Pu_D2D_fog,2017_Yang_DEBTS,2017_Mao_MobileEdge,2015_Kwak_JSAC_DREAM},
the Lyapunov optimization method was applied to transform the challenging stochastic
optimization problem to a sequential decision problem.

Note that all the above literatures assumed perfect knowledge about the system
parameters, e.g. the precise relationship among latency, energy consumption, and
computational resources. However, in practice, the models may be too complicated
to be modeled accurately and it is difficult to learn all the system parameters.
For example, the communication delay and the computation delay were modeled as
bandit feedbacks in \cite{2017_Chen_GG_BanditCVX}, which were only revealed for
the nodes that were queried. Without assuming particular system models and without
perfect knowledge about the system parameters, the tradeoff between learning the
system and pursuing the empirically best offloading strategy was investigated under
the bandit model in \cite{2018_Zhu_Task_Offload}. The exploration versus exploitation
tradeoff in \cite{2018_Zhu_Task_Offload} was addressed based on the multi-armed
bandit (MAB) framework, which has been extensively studied in statistics
\cite{1985_Berry_bandit,2012_Bubeck_Bandit,2002_Auer_UCB}.

In reality, it is hard to know the exact feedback rules of different fog nodes.
This reality motivates us to find one unified and analytical model involving several
features to approximate the behavior of fog nodes, e.g. whether they feel optimistic
about the incoming task or not. Additionally, it is unnecessary to estimate all the
parameters corresponding to every fog node since we just want to get optimistic
feedbacks after offloading each task. Clearly, there exists a tradeoff between
exploiting the empirically best node as often as possible and exploring other nodes
to find more profitable actions. In this paper, we model the
overall performance of offloading each task as one bit. This one-bit information
is fed back to the task node after completing the task. The value of the feedback
bit is modeled as a random variable and is related to some particular features of the
task and the node processing the task. We endeavor to make online decisions to
maximize the long-term performance of task offloading with these probabilistic
feedbacks. Our main contributions are summarized as follows. First, we apply a bandit
learning method to approximate the behavior of the concerned fog nodes and
model the node uncertainty with a logit model, which is more practical than the
previous ones, e.g.
\cite{2017_Yang_MEETS,2017_You_MobileEdge,2015_Kwak_JSAC_DREAM,2017_Mao_MobileEdge}.
Second, we extend the algorithm proposed in \cite{2016_Root} to make it suitable
to learn the feature weights of different fog nodes. We further analyze our proposed
algorithm and establish the corresponding performance guarantee for our extension.

The rest of this paper is organized as follows.
Section \ref{System_model} introduces the system model and assumptions.
Section \ref{Formulation} describes our algorithm and the related performance
guarantees. Numerical results are provided in section \ref{Simulation} and
Section \ref{Conclusions} concludes the paper.

\vspace{3pt}

\noindent{\it Notations}:
Notations $\bm A^\top$, $|\mathcal{A}|$, $\| \bm{x} \|_A^2$, $\Pr[A]$, and
$\mathbb E[A]$ stand for the matrix transpose, the cardinality of the
set $\mathcal{A}$, the norm with respect to $\bm A$, i.e. $\bm{x}^\top \bm{A}\bm{x}$
for a positive definite matrix $\bm{A}\in \mathbb R^{d \times d}$,
the probability of event $A$, and the expectation of a random variable $A$.
Indicator function $\mathds{1}\{\cdot\}$ takes the value of $1$ ($0$) when the specified condition is met (otherwise).

\begin{figure}[t]
\centering
\includegraphics[width = 0.43\textwidth]{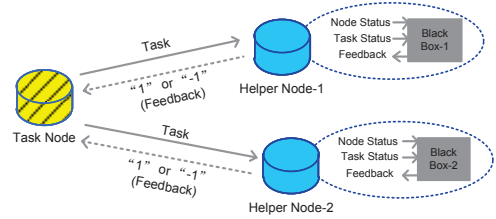}
\vspace{-6mm}
\caption{A task offloading example.
The task node offloads a task to a helper node and immediately receives a one-bit
feedback representing ``happy'' or ``unhappy''. The feedback is determined jointly
by the node status and the task details.}
\label{Fig:Fog_topology}
\vspace{-5mm}
\end{figure}

\vspace{-4mm}
\section{System Model} \label{System_model}
\vspace{-2mm}

We consider the task offloading problem in a network including $K$ fog nodes,
i.e. one task node and $(K-1)$ helper nodes. See Fig.~\ref{Fig:Fog_topology} for
an example. Define the set of fog nodes as
\vspace{-1mm}
\begin{equation}
\mathcal I := \{ \underbrace{1,2,\cdots,K-1}_{\rm helper~nodes},
\underbrace{K}_{\rm task~node} \}.
\vspace{-2mm}
\end{equation}
In each time slot, the task node generates one task and intelligently chooses
one fog node to offload this task. The helper nodes can also generate tasks
occasionally. In this paper, we focus on the offloading decisions at the task node.
We assume the tasks generated at the helper nodes are processed locally.
Further, we assume all the incoming tasks at each node are cached and executed in
a first-input first-output (FIFO) fashion.

In time slot-$t$, the task node offloads a task to a particular node-$I_t$ and
receives a one-bit feedback $y_{I_t}^{(t)}$. This one-bit feedback $y_{I_t}^{(t)}$
indicates whether the helper node feels optimistic about the current task.
Without loss of generality, we use $y_{I_t}^{(t)} = 1 \ (-1)$ to denote the node is
happy (unhappy). In this paper, we assume the feedback is delivered immediately after
receiving the offloaded task. As shown in Fig.~\ref{Fig:Fog_topology}, the feedback
is determined jointly by the task details and the node status.
To represent all the factors affecting the feedback $y_{i}^{(t)}$ from node-$i$,
we combine them into a feature vector
$\bm{x}_i^{(t)}$,
whose elements is a series of hypothetical features to depict the model.
The elements may include some real attributes, e.g. queue length, data length, task complexity, central processing unit (CPU) frequency, channel quality
information (CQI). In general cases, however, the features have no specific meaning.
Each element in $\bm{x}_i^{(t)}$ is normalized such that
$\|\bm x_i^{(t)}\|_2\le 1$. Note that different kinds of computing nodes certainly
have different preferences over various tasks, which can be reflected by applying
different weights to the features. Specifically, we can employ one weight vector
$\bm w_i$ and each element quantifies the weight associated with the corresponding
feature in $\bm{x}_i^{(t)}$. Similar to $\bm{x}_i^{(t)}$, we also normalize $\bm w_i$
such that $\|\bm w_i\|_2 \le 1$.

Note that it is hard to know the exact feedback rules of different fog nodes.
In this paper, we assume each node exploits the logit model, a commonly used binary
classifier \cite{2016_Root}, to evaluate the incoming tasks. Accordingly, the
probability of feeding back $y_i^{(t)}=1$ or $y_i^{(t)}=-1$ is given by
\vspace{-1mm}
\begin{equation}
\label{Eq:FeedbackProb}
\Pr\left[y_i^{(t)}=\pm 1 |\bm{x}_i^{(t)}\right] = \frac{1}{1 +
\exp \left(-y_i^{(t)} \bm{w}_i^\top \bm{x}_i^{(t)} \right) },
\vspace{-1mm}
\end{equation}
where the pair $(\bm x_i^{(t)},\bm w_i)$ is chosen from the following set
\begin{equation}
\label{Eq:Domain_D}
  \mathcal D_t := \{(\bm x_i^{(t)},\bm w_i), \forall i\in\mathcal I \}.
\end{equation}

\vspace{-4mm}
\section{Problem Formulation} \label{Formulation}
\vspace{-2mm}

Our goal is to maximize the long-term happiness metric. Consider a time range
$\mathcal T := \{1,2,\cdots,T\}$. The maximization of the long-term happiness metric
can be formulated as follows.
\begin{equation}\label{Eq:OriginalProblem}
\begin{split}
\underset{\{I_t\}}{\rm maximize} \quad   &
\underset{T\rightarrow\infty}{\rm lim} \
\frac{1}{T} \sum_{t\in\mathcal T} y_{I_t}^{(t)}  \\
{\rm subject \ to} \quad  &  (\ref{Eq:FeedbackProb}), \   {I_t}
\in\mathcal I, \forall t\in\mathcal T.
\end{split}
\end{equation}
There are two difficulties in (\ref{Eq:OriginalProblem}). First, the weight vector
$\bm w_i^{(t)}$ is unknown to the task node. Furthermore, the offloading decision is
made at the beginning of each time slot. Thus it is necessary to learn the weight
vectors along with making the task offloading decisions. To deal with the latter
one, we turn to solve the following problem as an alternative.
\begin{equation}
\label{Eq:RelaxedProblem}
\begin{split}
    \underset{I_t}{\rm maximize} \quad  &   \mathbb E [y_{I_t}^{(t)}]\\
    {\rm subject \ to} \quad  &  (\ref{Eq:FeedbackProb}), \   {I_t}\in\mathcal I.
\end{split}
\end{equation}
Although the above problem is not exactly the same as the original one in
(\ref{Eq:OriginalProblem}), it is one common approach and was adopted in \cite{2017_Mao_MobileEdge,2017_Yang_DEBTS,2017_Pu_D2D_fog,2018_Zhu_Task_Offload}.
Meanwhile, under the stochastic framework \cite{2012_Bubeck_Bandit}, it is more
natural to focus on the expectation, i.e. $\mathbb E [y_{I_t}^{(t)}]$.
Note the expected happiness metric of each arm has to be estimated based on the
historical feedback. There is thus an exploration-exploitation tradeoff in
(\ref{Eq:RelaxedProblem}). On the one hand, the task node tends to choose the
best node according to the historical information. On the other hand, trying offloading
to unfamiliar nodes may bring task node extra rewards. Plenty of works have been done
to deal with this kind of exploration-exploitation tradeoff problem under the MAB
framework \cite{2018_Zhu_Task_Offload,2002_Auer_UCB,1985_Berry_bandit,
2012_Bubeck_Bandit,2016_Root}. In the rest of the paper, we also address this
tradeoff through the bandit methods.

\vspace{-3mm}
\section{Online task offloading}\label{Algorithms}
\vspace{-2mm}

\vspace{-0mm}
\subsection{Task Offloading with One-bit Feedback}
\vspace{-1mm}

This exploration-exploitation tradeoff can be solved with a stationary multi-armed
bandit (MAB) model, where each node can be viewed as one arm. Offloading one task
is like testing one arm and the task node makes decisions according to all the
feedbacks it has received.
Given the first $T$ observations of the feedbacks, i.e. $y_{I_t}^{(t)}, t =
1,2,\cdots,T$, the weight vector of each fog node can be approximated by its maximum
likelihood estimate as follows.
\vspace{-2mm}
\begin{equation}
\bar{\bm w}_i^{(T)} = \arg\max_{\|\bm w\|\le 1} \frac{1}{T} \sum_{t=1}^T
f_i^{(t)}(\bm w), \forall i\in\mathcal I,
\end{equation}
where the log likelihood function is defined based on (\ref{Eq:FeedbackProb}),
\begin{equation}
  f_i^{(t)}(\bm w) = - \log \left( 1+\exp(-y^{(t)}_i {\bm w}^{\top} {\bm x_i} ) \right) \mathds 1\{I_t = i\}.
\end{equation}
Clearly, this approach needs to optimize over all the historical feedbacks, which is
not scalable. To admits online updating, we refer to \cite{2016_Root} and
propose an approximate sequential MLE solution as
\begin{equation}
  \begin{split}
  \label{Eq:W updating function}
    \bar {\bm w}_i^{(t+1)}
    & =  {\arg \max_{\| \bm{w} \|_2 \leq 1}} - \frac{\| \bm{w}- \bar {\bm w}_i^{(t)} \|_{\bm Z_i^{(t)}}}{2} \\
    & +  (\bm{w}- \bar {\bm w}_i^{(t)})^\top \nabla f_i^{(t)}(\bar {\bm w}_i^{(t)}) \mathds 1\{I_t = i\},
  \end{split}
\end{equation}
where
\begin{equation}\label{Eq:UpdateZ}
   \bm Z_i^{(t+1)} = \bm Z_i^{(t)} +  \frac{\beta}{2} {\bm x}_i^{(t)} ({\bm x}_i^{(t)})^{\top} \mathds 1\{I_t = i\}.
\end{equation}
The term $\| \bm{w}- \bar {\bm w}_i^{(t)} \|_{\bm Z_i^{(t)}}$ in (\ref{Eq:W updating function}) is an exploration bonus.
Specifically, if one fog node is explored deficiently, the restriction given by $\bm Z_i^{(t)}$ on the exploration bonus term is relatively loose.
Thus the wider range of exploration of this node is more recommended.

As indicated in (\ref{Eq:RelaxedProblem}), our goal is to maximize the expectation
of instantaneous happiness metric, which is positively correlated to the probability
of $y_{I_t}^{(t)} = 1$. Additionally, the metric is positively correlated to
$\bm w_i^\top \bm x_i$ as well. In time slot-$t$, the task node then chooses one fog
node to offload based on the feature $\bm x_i^{(t)}$ by solving the following
optimization problem:
\begin{equation}\label{Eq:Choose_x_w}
  ({\bm x}_{I_t}^{(t)} , \hat{\bm w}_{I_t}) = \arg\max_{(\bm x,\bm w)\in \bar{\mathcal D}_t }\bm w^\top \bm x.
\vspace{-2mm}
\end{equation}

Note $\hat{\bm w}_{I_t}$ is just a temporal variable that does not engage in the
updates of any variables. Essentially, we are only interested in the index of the node,
i.e. $I_t$. The domain $\bar{\mathcal D}_t$ is defined as
\begin{equation}
  \bar{\mathcal D}_t = \bigcup_{i\in\mathcal I} \{(\bm x,\bm w)| \bm x = \bm x_i^{(t)},\bm w \in\mathcal W_i^{(t)}  \},
\end{equation}
and $\mathcal W_i^{(t)}$ denotes the feasible region of the estimated weights,
which is a ball centered at $\bar {\bm w}_i^{(t)}$.
Specifically, the ball is characterized as
\begin{equation}\label{Eq:feasible_w}
  \mathcal W_i^{(t)} := \{ \bm w | \| \bar{\bm w}_i^{(t)} - {\bm w} \|_{{\bm Z}^{(t)}_i}^2 \le {\gamma^{(t)}_i}\}.
\end{equation}
The benefit of the exploration will be further explained in section \ref{Guarantees}.
Note ${\gamma^{(t)}_i}$ is an important parameter, the value of which determines the
performance of our proposed algorithm. Details about ${\gamma^{(t)}_i}$ and the
corresponding theoretical guarantees will be discussed later in Section
\ref{Guarantees}. Based on the feasible region of $\hat{\bm w}_{I_t}$ defined in
(\ref{Eq:feasible_w}), we can identify the node index $I_t$ in (\ref{Eq:Choose_x_w})
as follows.
\vspace{-1mm}
\begin{equation}\label{Eq:SelArm}
\begin{split}
{I_t}
& = \arg\max_{i\in\mathcal I} \ \left(\max_{\| \bar{\bm w}_i^{(t)} -
{\bm w} \|_{{\bm Z}^{(t)}_i}^2 \le {\gamma^{(t)}_i}}  \bm w^\top \bm x_i^{(t)}\right) \\
& = \arg\max_{i\in\mathcal I} \ \left(   (\bar{\bm w}_i^{(t)})^\top \bm x_i^{(t)}
- \min_{\| \bm z\|_{2}^2 \le {\gamma^{(t)}_i}} [(\sqrt{{\bm Z}^{(t)}_i})^{-1}
{\bm z}]^\top \bm x_i^{(t)} \right) \\
& = \arg\max_{i\in\mathcal I} \ \left(   (\bar{\bm w}_i^{(t)})^\top \bm x_i^{(t)}
+ \sqrt{\gamma^{(t)}_i} \|\bm x_i^{(t)}\|_{\left(\bm Z_i^{(t)}\right)^{-1}} \right).
\end{split}
\end{equation}
The proposed strategy, i.e. Task Offloading with One-bit Feedback (TOOF), is
summarized in Algorithm \ref{Alg:TOOF}.
Referring to (\ref{Eq:W updating function}), (\ref{Eq:UpdateZ}), and (\ref{Eq:SelArm}), our updating strategy relies on the latest feedback rather than the accumulated history information.
Thus, it can be executed in an online fashion with remarkably low complexity.
It's worth mentioning that the TOOF resorts to a UCB-type algorithm and
the deciding rule of $I_t$ in (\ref{Eq:SelArm}) functions as the upper confidence bound as in \cite{2002_Auer_UCB}.
\begin{algorithm}[t]
\caption{Task Offloading with One-bit Feedback (TOOF)}
\begin{algorithmic}[1]
\State \textbf{Initialization} $\lambda =1$, $\bm Z_i^{(1)} = \lambda \bm I$, $\bm w_i^{(1)} = \bm 0$, $\forall i\in\mathcal I$;
\State \textbf{for} $t = 1,2,\cdots $ \textbf{do}
\State \quad \textbf{if} $t \le K$
\State \quad \quad Observe feature vector $\bm{x}_i^{(t)}$; select node-$I_t = t$;
\State \quad \textbf{else if} $t > K$
\State \quad \quad Observe feature vector $\bm{x}_i^{(t)}$, select node-$I_t$ as (\ref{Eq:SelArm});
\State \quad \textbf{end if}
\State \quad Transmit task-$t$ to node-$I_t$, observe $y_{I_t}^{(t)} \in \{ \pm 1 \}$;
\State \quad Update $\bm{w}_i^{(t+1)}$ and $\bm Z_i^{(t+1)}$ as (\ref{Eq:W updating function}) and (\ref{Eq:UpdateZ}), $\forall i\in\mathcal I$;
\State \textbf{end for}
\end{algorithmic}
\label{Alg:TOOF}
\end{algorithm}

\vspace{-2mm}
\subsection{Theoretical Guarantees}\label{Guarantees}
\vspace{-1mm}

We provide theoretical analyses for our proposed algorithm when the actual feedback
model\footnote{The actual model may be arbitrary. The analysis of model mismatching
is left for our future works.} is the same as the one in (\ref{Eq:FeedbackProb}).
The convergence of $\bm w_i$ is provided in Proposition \ref{Pro:Est_W}.
Note the proof is similar to the one for Theorem 1 in \cite{2016_Root}.

\begin{pro}\label{Pro:Est_W}
With a probability at least $(1-\delta)$, we have
\begin{equation}
\label{Eq:Bound_w}
\| \bar{\bm w}_i^{(t+1)} - {\bm w}_i \|_{{\bm Z}^{(t+1)}_i}^2 \le {\gamma^{(t+1)}_i},
\forall t>0 ,
\end{equation}
where $\delta$ is a control parameter, and
\begin{equation}
\label{gamma:update}
    \gamma^{(t+1)}_i = \left[ 8 + \left( \frac{8}{\beta} + \frac{16}{3}  \right)\tau_t + \frac{2}{\beta} \log\frac{\det (\bm Z^{(t+1)}_i)}{\det (\bm Z^{(1)}_i)} \right] + \lambda,
  \end{equation}
  \begin{equation}
    \tau_t = \log\left( \frac{2\lceil 2\log_2 t \rceil t^2 }{\delta} \right), \ \ \beta = \frac{1}{2(1+\exp(1))}.
\end{equation}
\end{pro}

Proposition \ref{Pro:Est_W} indicates that the width of the confidence region, i.e.
$\gamma^{(t+1)}_i$, is in the order of $O(\sqrt{d\log t})$, where $d$ is a particular
constant. By carefully choosing the value of $\delta$, we can say that the weight
vector ${\bm w}_i$ is in $\mathcal W_i^{(t)}$ with a sufficiently high probability.
If the weight vector of each node is perfectly observed, the task node can
pick a node with the maximal probability of positive feedback. Thus, we define the
optimal node in time slot-$t$ as node-$I^*_t$ such that
\begin{equation}\label{Eq:ChooseOpt}
({\bm x}_{I_t^*}^{(t)}, {\bm w}_{I_t^*} ) = \arg\max_{(\bm x,\bm w)\in\mathcal D_t}
\bm w^\top \bm x,
\end{equation}
where the domain $\mathcal D_t$ is defined in (\ref{Eq:Domain_D}).
Accordingly, the instantaneous regret function could be written as follows.

\begin{equation} \label{Regret_function}
  r_t = \left(\bm w_{I_t^*}^\top \bm{x}_{I_t^*} - \bm w_{I_t}^\top \bm{x}_{I_t}^{(t)}\right).
\end{equation}

The upper bound on the regret is given in proposition \ref{Pro:r_t}.

\begin{pro}\label{Pro:r_t}
With a probability at least $(1-\delta)$, the average regret,
i.e. $R(T):= \frac{1}{T} \sum_{t=1}^T r_t$ is upper-bounded as
\begin{equation}
\begin{split}
R(T) \le  4\sqrt{ \frac{\gamma^{(T)}}{\beta T}  \sum_{i=1}^K \log\frac{\det(\bm Z_{i}^{(T)})}{\det(\bm Z_{i}^{(1)})}},
\end{split}
\end{equation}
where $\gamma^{(T)}=\max_{i\in\mathcal I} \gamma_i^{(T)}$.
\end{pro}

This proposition implies the average regret approaches to zero as the time goes to
infinity with overwhelming probability. Additionally, the upper bound is in the order
of $O(\sqrt{(\log t)/t})$. The proof outline can be found in Appendix.

\vspace{-4mm}
\section{Numerical Results} \label{Simulation}
\vspace{-2mm}

In this section, we examine the performance of our algorithm by testing $T = 2000$
tasks and compare the performance with other algorithms. The tasks are allocated to
$ K = 10$ fog nodes on demand. Besides, we assume that data length uniformly
distributed within $[1,15]$KB. For each task, $\bm x_i$ consists of five features.
In particular, features including ``task length'', ``task complexity'', and ``
queue length'' are negatively correlated to the happiness of a node. Meanwhile,
features including ``CPU frequency'' and ``CQI'' are positively correlated.
The parameter $\lambda$ is introduced to make
sure that $Z_t$ is invertible and barely affects the performance of our algorithm.
Hence, we simply choose $\lambda = 1$ according to \cite{2016_Root}.
The parameter $\gamma^{(t)}_i$ is tuned to be $c\log \frac{\det(\bm Z_{i}^{(t)})}{\det(\bm Z_{i}^{(1)})}$
according to (\ref{gamma:update}) where $c=0.01$. It is worth noting that the value
of $\gamma^{(t)}_i$ has the same order as that in (\ref{gamma:update}) instead of
the exact value. This is due to the fact that the $\gamma^{(t)}_i$ in
(\ref{gamma:update}) only provides an upper bound on the estimation error of
$\bm w_i$, which may not be tight enough in terms of the aforementioned constant $d$
in Proposition \ref{Pro:Est_W}.

\begin{figure}[t]
\centering
\includegraphics[width = 0.43\textwidth]{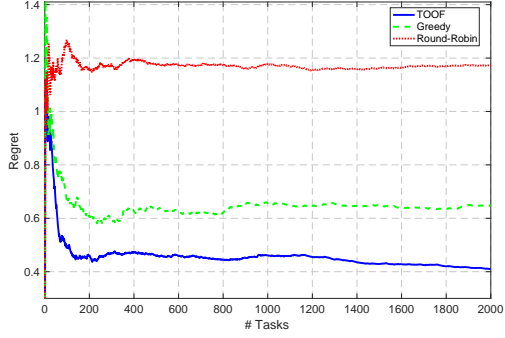}
\vspace{-10pt}
\caption{Average regret versus time. The regret is calculated as $R(T) := \frac{1}{T}\sum_{t=1}^T r_t$.}
\label{Fig:Regret}
\vspace{-10pt}
\end{figure}

\begin{figure}[t]
\centering
\includegraphics[width = 0.43\textwidth]{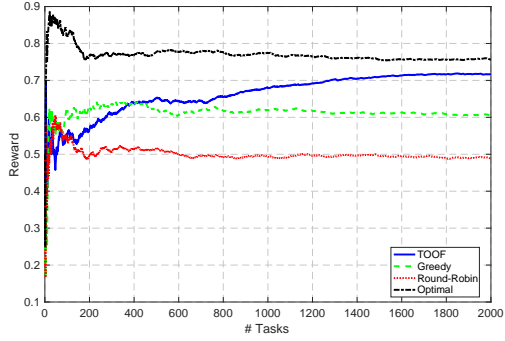}
\vspace{-10pt}
\caption{Average reward versus time. The reward is calculated as $\tilde R(T) := \frac{1}{T}\sum_{t=1}^T \mathds 1\{y_{I_t}^{(t)}=1\}$.}
\label{Fig:Happy}
\vspace{-10pt}
\end{figure}

In Fig.~\ref{Fig:Regret}, we compare the performance of the TOOF algorithm with
{\it Round-Robin} and {\it Greedy}. In the round-robin algorithm, nodes are chosen
in a cyclic sequence regardless of their current states.
In the greedy algorithm, the task node chooses a helper node in each time slot under
the same rule as TOOF, but $Z_t$ stays the same over time.
It means that each single element of the estimated weight vector $\bar {\bm w}_i$ is
updated in the same pace. Clearly, Fig.~\ref{Fig:Regret} indicates that our
proposed TOOF algorithm shows the tendency of converging to zero. Besides, the TOOF
algorithm achieves much lower regret than the other two algorithms.

The superior performance of our proposed scheme is also shown in Fig.~\ref{Fig:Happy}.
Comparing with (\ref{Eq:OriginalProblem}), we find that the reward defined in
Fig.~\ref{Fig:Happy} is also a happiness metric by denoting happy (unhappy) by
$y_{I_t}^{(t)}=1 (0)$. In Fig.~\ref{Fig:Happy}, {\it Optimal} shows
the performance in the case of perfect knowledge, where the node is chosen as
(\ref{Eq:ChooseOpt}). Note that the regret of {\it Optimal} is a zero value function.
Our algorithm begins to show its superiority to the greedy
algorithm since time slot-$400$ and keeps widening the gap. Fig.~\ref{Fig:Happy} also
illustrates that the reward obtained via the TOOF algorithm approaches the optimal one.
This shows that, with the increment of the number of tasks, our TOOF algorithm is
capable of dealing with the tradeoff between learning system parameters and getting
a high immediate reward.

\vspace{-2mm}
\section{Conclusions} \label{Conclusions}
\vspace{-2mm}

In this paper, we have investigated an efficient task offloading strategy with one-bit
feedback and have established the corresponding performance guarantee.
Without knowing weight vectors of the helper nodes, with the probabilistic feedbacks,
a multi-armed bandit framework has been formulated. Under the framework, we have
proposed an efficient TOOF algorithm basing on the UCB policy.
We have also proven that the upper bound of the average regret function is in the
order of $O(\sqrt{(\log t)/t})$. Numerical simulations also demonstrate that
our TOOF algorithm is able to obtain superior performance in an online fashion.

\vspace{-3mm}
\section{Appendix} \label{Appendix}
\vspace{-2mm}

The following inequality always holds due to that ${\bm w}_{I_t}$ and ${\bm x}_{I_t}^{(t)}$ have been normalized unitary:
\vspace{-2mm}
\begin{equation}
  \begin{split}
    r_t =  & {\bm w}_{I_t^*}^{\top} {\bm x}_{I_t^*}^{(t)} - {\bm w}_{I_t}^{\top} {\bm x}_{I_t}^{(t)} \\
        =  & {\bm w}_{I_t^*}^{\top} ({\bm x}_{I_t^*}^{(t)} - {\bm x}_{I_t}^{(t)}) + ({\bm w}_{I_t^*} - {\bm w}_{I_t})^\top {\bm x}_{I_t}^{(t)}  \le 4.
  \end{split}
\end{equation}
On the other hand, with a probability at least $(1-\delta)$,
the instantaneous regret $r_t$ can be upper-bounded as follows.
\vspace{-2mm}
\begin{equation}
\nonumber
\begin{split}
r_t &= {\bm w}_{I_t^*}^{\top} {\bm x}_{I_t^*}^{(t)} - {\bm w}_{I_t}^{\top} {\bm x}_{I_t}^{(t)} \\
&\le  (\hat{\bm w}_{I_t}^{(t)} - \bar{\bm w}_{I_t}^{(t)})^\top  {\bm x}_{I_t}^{(t)} 
+  (\bar{\bm w}_{I_t}^{(t)} - {\bm w}_{I_t})^{\top} {\bm x}_{I_t}^{(t)} \\
& \overset{(a)}{\le} \left(\|\hat{\bm w}_{I_t}^{(t)} - 
\bar{\bm w}_{I_t}^{(t)}\|_{\bm Z^{(t)}_i} + \|\bar{\bm w}_{I_t}^{(t)} - 
{\bm w}_{I_t}\|_{\bm Z^{(t)}_i}\right) \|{\bm x}_{I_t}^{(t)}\|_{(\bm Z^{(t)}_i)^{-1}}
\\
&\overset{(b)}{\le} 2\sqrt{\gamma_i^{(t)}} \|{\bm x}_{I_t}^{(t)}\|_{(\bm Z^{(t)}_{I_t})^{-1}}.
\end{split}
\end{equation}
where $(a)$ holds due to the Cauchy–Schwarz inequality, and $(b)$ holds with a probability at least $(1-\delta)$ referring to Proposition 1. Then the total
regret can be upper-bounded by
\begin{equation}
\label{Eq:regret_bound_1}
\begin{split}
& \sum_{t=1}^T r_t = \sum_{t=1}^T \left( {\bm w}_{I_t^*}^{\top} {\bm x}_{I_t^*}^{(t)} 
- {\bm w}_{I_t}^{\top} {\bm x}_{I_t}^{(t)} \right) \\
& \le \sum_{t=1}^T \min \left( 2\sqrt{\gamma^{(t)}} \| \bm x_{I_t}^{(t)} \|_{(\bm Z_{I_t}^{(t)})^{-1}} , 4 \right) \\
& \le \sqrt{ \frac{8\gamma^{(T)}}{\beta} }  \max \left(1, \sqrt{2\beta} R \right)\sum_{t=1}^T \min \left( 
\sqrt{\frac{\beta}{2}} \| \bm x_{I_t}^{(t)} \|_{(\bm Z_{I_t}^{(t)})^{-1}} , 1 
\right) \\
& \le \sqrt{ \frac{8\gamma^{(T)}T}{\beta} }  \sqrt{\sum_{t=1}^T \min 
\left( {\frac{\beta}{2}} \| \bm x_{I_t}^{(t)} \|^2_{(\bm Z_{I_t}^{(t)})^{-1}} , 
1 \right)}. \\
\end{split}
\vspace{-3mm}
\end{equation}
Similar to the result from Lemma 11 in \cite{2011_lemma}, we have
\vspace{-1mm}
\begin{equation}
\begin{split}
& \det\left(\bm Z_{i}^{(T+1)}\right) \\
& = \det\left( \bm Z_{i}^{(T)} + \frac{\beta}{2} \bm x_{i}^{(T)} 
(\bm x_{i}^{(T)})^\top \mathds 1\{I_T = i\} \right) \\
& = \det\left(\bm Z_{i}^{(T)}\right) \left(1 + \frac{\beta}{2} \| 
\bm x_{i}^{(T)} \|^2_{(\bm Z_{i}^{(T)})^{-1}} \mathds 1\{I_T = i\}\right) \\
& = \det\left(\bm Z_{i}^{(1)}\right) \prod_{t=1}^T \left(1 + \frac{\beta}{2} 
\|\bm x_{i}^{(t)} \|^2_{(\bm Z_{i}^{(t)})^{-1}} \mathds 1\{I_t = i\}\right).
\end{split}
\end{equation}
\vspace{-1mm}
Thus we have
\vspace{-0mm}
\begin{equation}
\begin{split}
&\sum_{t=1}^T \min \left( {\frac{\beta}{2}} \| \bm x_{I_t}^{(t)} \|^2_{(\bm Z_{I_t}^{(t)})^{-1}} , 1 \right) \\
&\le 2 \sum_{t=1}^T \log \left( 1 + {\frac{\beta}{2}} \| \bm x_{I_t}^{(t)} 
\|^2_{(\bm Z_{I_t}^{(t)})^{-1}} \right) \\
&=  2 \sum_{t=1}^T \sum_{i=1}^K \log \left( 1 + {\frac{\beta}{2}} \|
 \bm x_{i}^{(t)} \|^2_{(\bm Z_{i}^{(t)})^{-1}} \mathds 1\{I_t=i\} \right) \\
&=  2 \sum_{i=1}^K \log\frac{\det\left(\bm Z_{i}^{(T)}\right)}{\det
\left(\bm Z_{i}^{(1)}\right)}.
  \end{split}
\end{equation}
Taking this result to (\ref{Eq:regret_bound_1}) yields
\begin{equation}
  \sum_{t=1}^T r_t \le 4\sqrt{ \frac{\gamma^{(T)}T}{\beta}  \sum_{i=1}^K \log\frac{\det(\bm Z_{i}^{(T)})}{\det(\bm Z_{i}^{(1)})}}. \\
\end{equation}

\end{document}